\documentclass[sigconf]{acmart-me}

\usepackage{booktabs}
\usepackage{url}
\usepackage{color}
\usepackage{enumitem}
\setcopyright{rightsretained}

\acmDOI{}

\acmISBN{}

\acmConference[MediaEval'21]{Multimedia Evaluation Workshop}{December 13-15 2021}{Online}
\acmYear{2021}
\copyrightyear{}

\acmPrice{}

\begin{document}
\title{Spatio-Temporal CNN baseline method for the Sports Video Task of MediaEval 2021 benchmark}

\author{\vspace{-15pt}Pierre-Etienne Martin}

\affiliation{CCP Department, Max Planck Institute for Evolutionary Anthropology, D-04103 Leipzig, Germany}

\email{pierre_etienne_martin@eva.mpg.de}

\renewcommand{\shorttitle}{Sports Video Task}

\begin{abstract}
This paper presents the baseline method proposed for the Sports Video task part of the MediaEval 2021 benchmark. This task proposes a stroke detection and a stroke classification subtasks. This baseline addresses both subtasks.
The spatio-temporal CNN architecture and the training process of the model are tailored according to the addressed subtask.
The method has the purpose of helping the participants to solve the task and is not meant to reach state-of-the-art performance. Still, for the detection task, the baseline is performing better than the other participants, which stresses the difficulty of such a task.
\vspace{-5pt}
\end{abstract}

\begin{teaserfigure}
\vspace{-16pt}
    \includegraphics[width=.9\linewidth]{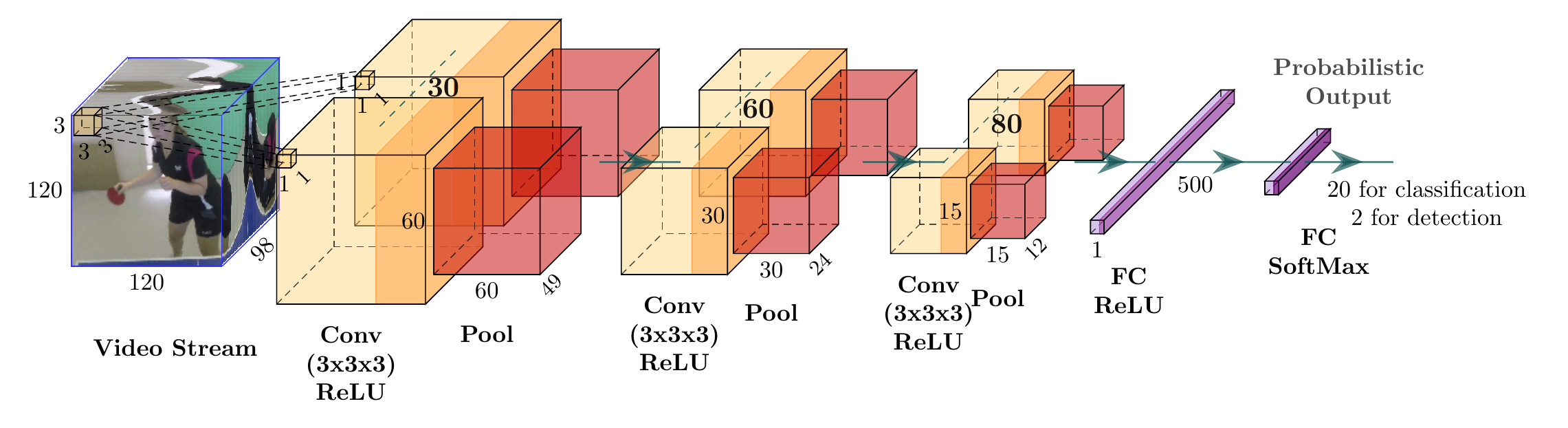}
    \vspace{-16pt}
    \caption{Spatio-Temporal CNN architecture for Stroke Classification and Detection.}
    \label{fig:dataset}
  \label{fig:teaser}
\end{teaserfigure}

\maketitle
\vspace{-8pt}
\section{Introduction}
\label{sec:intro}

Most recent action detection and classification methods developed in the literature have been using deep learning approaches and high-dimensional spaces. In the domain of image classification, a specific kind of Neural Network has become very popular: the Convolutional Neural Networks (CNNs). Since the breakthrough at the $2012$ ImageNet Challenge, CNNs have demonstrated a great improvement for image classification.
\par
For video applications in general and action recognition in particular, the first models proposed were a direct extension of image classification methods~\cite{Hakan:2018, NN:SimonyanTwoStream:2014} using 2D convolutions. However, to better capture the temporal information proper to video content, the use of 3D convolutions has emerged~\citep{NN:3DCNN_first:2007,NN:3DCNN:2017}. One can also consider temporal information using the motion extracted from successive frames, such as the optical flow. The latest can be used i) as a single modality or in parallel with the RGB information~\citep{NN:TwoStream3D:2018, NN:I3DCarreira:2017, NN:SimonyanTwoStream:2014, TT:OFSingularitiesForStrokeClassification:2019}; or ii) to train a network for extracting motion features to perform classification at a later stage~\cite{NN:3DCNNMimiciMotionForCLassification:2019}. These methods also raise the question of how to fuse the different modalities~\citep{NN:TwoStreamFusion:2016, PeThesis2020}. In~\cite{Pose:ActionRecognitionAndPoseEstimation:2018, Pose:LCRNet++:2020}, the estimated pose is used jointly with these two modalities to perform action classification. In~\cite{Pe:3stream:2021} all the three modalities are used and fused in order to perform stroke classification.
\par
As part of the task organization, for the first time since the beginning of the Sports Video task (in 2019~\citep{PeMETask:2019}), we decided to provide a baseline to alleviate minor aspects of the task, such as video and xml processing; and help the participants in their submission. The baseline method uses a 3D CNN inspired from~\cite{PeMEWork:2019,PeCBMI:2018}. We adjusted the method to answer both proposed subtasks of this year's edition~\cite{mediaeval/Martin21/task}: stroke detection and stroke classification from videos of the \texttt{TTStroke-21} corpus. The implementation of the method is available publicly on Github\footnote{\url{https://github.com/ccp-eva/SportTaskME21}}.

\vspace{-5pt}
\section{Method}
\label{sec:dataset}

In order to perform classification and detection, we consider the model architecture presented in Fig.~\ref{fig:teaser}. For each subtask, a distinct model has been trained on the train set. We train both using a stochastic gradient approach with a Nesterov momentum of 0.5~\cite{Deep:Nesterov1}, a weight decay of 0.005~\citep{Deep:weight_decay} and a constant learning rate of 0.0001. Both models are trained over 500 epochs. The objective function is the cross-entropy loss of the output processed by the softmax function (eq. \ref{eq:CrossEntropyLoss}) summing over the batch:
\vspace{-2pt}
\begin{equation}
    \label{eq:CrossEntropyLoss}
    \mathcal{L}(y,class) = -log(\dfrac{exp(y_{class}')}{\sum_i^Nexp(y_i)})
\end{equation}

At each epoch, the model is validated on the validation set. The model performing the best on this set is saved and then evaluated on the test set. The model is fed with the video frames resized to $120\times120$ and staked successively in cuboids of length 98, representing approximately 0.82 seconds.
\par
For the detection task, we inferred \textit{Non-stroke} segments from the annotated \textit{Stroke} segments. We considered only segments between two consecutive strokes greater than 200 frames. Such a segment is divided in successive blocks of 200 frames, non overlapping, and added has a negative sample for training the model. The split using 200 frames allows a correct number of negative samples: from the 783 train and 234 validation segments, we inferred respectively for each set 1196 and 260 negative segments. No negative segments have been inferred from the test set. Stroke detection is tackled as a classification task by considering two classes: \textit{Stroke} on \textit{Non-stroke}. From the test set, which has no temporal boundaries, we created window proposals of length 150 every 150 frames for all the videos. This size was chosen empirically and meant to be revised to achieve good performance. For the classification task, all the classes were not represented in the dataset but we still consider all the 20 possible stroke classes.
\par
To train the model, we inputted the RGB cuboids composed of the successive frames from the \textit{starting frame} of the considered segment. The desired output is the class vector summing to one and binary at training time. Its length is the number of considered classes: 2 for detection and 20 for classification. Each element represents the probability of belonging to a class. During inference, we follow a similar procedure, and the class decision is the \textit{argmax} of the output vector.
\vspace{-7pt}
\section{Results}

This section presents the results per subtask according to the metrics presented in~\cite{mediaeval/Martin21/task}.
\vspace{-7pt}
\subsection{Subtask 1 - Stroke Detection}

The detection subtask was tackled as a classification task, considering the strokes and non-strokes samples. After 500 epochs, the model reached 98.3\% and 75.7\% of accuracy, respectively, on the train and validation sets. On the test set, the model is evaluated using the mAP metric. This metric takes into account the number of actions detected and their overlapping with the ground truth. The baseline achieves an mAP of 0.0173, which the two participants of this subtask did not outperform.
\par
Runs are also evaluated using a global IoU that considers only the frame-wise overlap of the detected strokes with the ground truth annotations. The number of strokes detected is no longer taken into account in the evaluation. The baseline achieves a Global IoU of 0.144, which was outperformed by one participant.
\par
The method’s performance is quite low due to the method being relatively simple. It also relies on a straightforward and non-efficient window proposal to segment the strokes without fusing the output decision. Indeed, two consecutive windows, part of the same stroke and classified as strokes, will be classified as two different strokes and not a single one, and will therefore have an impact on the mAP metric. The method can easily be improved by considering better proposals and fusing the output decisions.
\vspace{-7pt}
\subsection{Subtask 2 - Stroke Classification}

\begin{figure}
    \begin{tabular}{cc}
        \includegraphics[width=.45\linewidth]{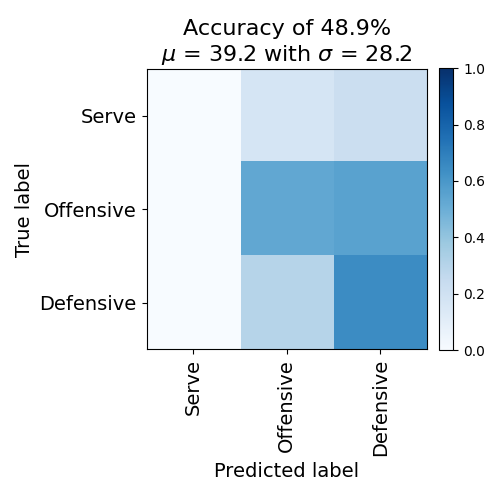} &
        \includegraphics[width=.45\linewidth]{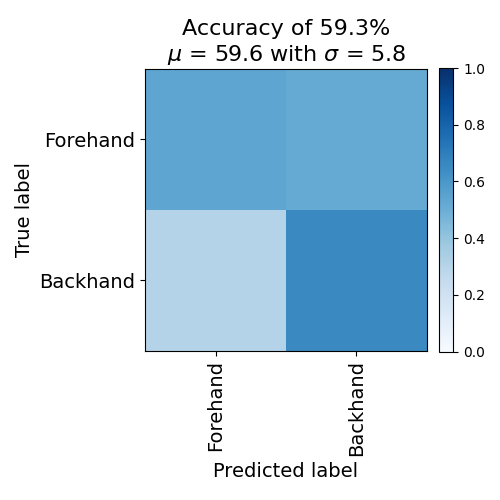} \\
        \textbf{a.} Type & \textbf{b.} Hand-side 
    \end{tabular}
    \vspace{-10pt}
    \caption{Confusion matrices with higher level categories.}
    \vspace{-20pt}
    \label{fig:confless}
\end{figure}

The results for the stroke classification subtask on the test set are reported in the table~\ref{Table:Acc}. This table is divided into different sections for considering different refined classifications. After training, the model reached only 25.2\% and 28.9\% of accuracy, respectively, on the train and validation sets.
\begin{itemize}
    \item ``Global'' consider all the 20 classes
    \item ``Type'' consider only the type of the stroke: Defensive, Offensive or Service
    \item ``Hand-Side'' consider only Forehand and Backhand super-classes
    \item ``Type and Hand-Sided'' consider the intersection of the two last clusters leading to 6 classes.
\end{itemize}
The confusion matrix of the ``Type'' and ``Hand-Side'' are also depicted in Fig.~\ref{fig:confless} for further analysis.

\begin{table}
\caption{Baseline performance in term of accuracy (\%).}
\vspace{-8pt}
  \label{Table:Acc}
  \begin{tabular}{c|c|c|c}
    \toprule
    Global &Type and Hand-Sided & Type & Hand-Side\\
    \midrule
    20.4 & 33 & 48.9 & 59.3\\
  \bottomrule
\end{tabular}
\vspace{-10pt}
\end{table}

From table~\ref{Table:Acc}, we can state that the performance of the baseline, considering all classes, is limited. This may be improved by further analysis of the corpus and further training. Indeed, only 18 classes over the 20 possible were present in the corpus this year, which simplifies the complexity of the task and could have been taken into account in the model's design. Fig.~\ref{fig:confless}.a reveals that the services have not been learned at all, which is undoubtedly due to the input processing during training which considers only the 100 first frames and is therefore unable to capture features from these longer strokes. Finally, Fig.~\ref{fig:confless}.b underlines the main weakness of the model: being unable of distinguishing Forehand and Backhand strokes. The pipeline's method could consider higher level categories, following a cascade method, to improve the performance. Two of the three participants have outperformed by far the baseline performance~\cite{mediaeval21/transformer, mediaeval21/hcmus}.
\vspace{-5pt}
\section{Conclusion}
\label{sec:discussion}

This baseline intends to help the participants solving the Sports Video Task. The baseline performance remains limited, but its publicly available implementation allows the participants to not start from scratch. Many aspects of the method may be improved, such as the data processing: a spatial and temporal ROI may increase the performance. Similarly with the architecture of the model, which was kept very simple, or the training method that could have merged the train and validation sets before inferring on the test set.
\par
The detection subtask seems to be challenging. No participants were able to beat the baseline performance with regard to the mAP metric, which is the ranking metric. This subtask is new in the Sports Video Task, which also explains the low results obtained. However we believe much improvement can be obtained since our method has tackled it as a classification task. The window proposal is also very crude and can easily be improved. 
\par
The classification subtask has gathered more participants with, overall, more successful performance. This may be explained by the task's non-novelty in the history of the MediaEval benchmark and the more active investigation in this field.
\par
Next year we plan to gather ideas from this year's submissions to improve the baseline and give a more substantial base to the new participants joining the Sports Video Task.

\bibliographystyle{ACM-Reference-Format}
\bibliography{sigproc} 


\begin{thebibliography}{00}


\ifx \showCODEN    \undefined \def \showCODEN     #1{\unskip}     \fi
\ifx \showDOI      \undefined \def \showDOI       #1{#1}\fi
\ifx \showISBNx    \undefined \def \showISBNx     #1{\unskip}     \fi
\ifx \showISBNxiii \undefined \def \showISBNxiii  #1{\unskip}     \fi
\ifx \showISSN     \undefined \def \showISSN      #1{\unskip}     \fi
\ifx \showLCCN     \undefined \def \showLCCN      #1{\unskip}     \fi
\ifx \shownote     \undefined \def \shownote      #1{#1}          \fi
\ifx \showarticletitle \undefined \def \showarticletitle #1{#1}   \fi
\ifx \showURL      \undefined \def \showURL       {\relax}        \fi
\providecommand\bibfield[2]{#2}
\providecommand\bibinfo[2]{#2}
\providecommand\natexlab[1]{#1}
\providecommand\showeprint[2][]{arXiv:#2}

\bibitem[\protect\citeauthoryear{Bilen, Fernando, Gavves, and Vedaldi}{Bilen
  et~al\mbox{.}}{2018}]%
        {Hakan:2018}
\bibfield{author}{\bibinfo{person}{Hakan Bilen}, \bibinfo{person}{Basura
  Fernando}, \bibinfo{person}{Efstratios Gavves}, {and} \bibinfo{person}{Andrea
  Vedaldi}.} \bibinfo{year}{2018}\natexlab{}.
\newblock \showarticletitle{Action Recognition with Dynamic Image Networks}.
\newblock \bibinfo{journal}{{\em {IEEE} Trans. Pattern Anal. Mach. Intell.\/}}
  \bibinfo{volume}{40}, \bibinfo{number}{12} (\bibinfo{year}{2018}),
  \bibinfo{pages}{2799--2813}.
\newblock


\bibitem[\protect\citeauthoryear{Calandre, P{\'{e}}teri, and
  Mascarilla}{Calandre et~al\mbox{.}}{2019}]%
        {TT:OFSingularitiesForStrokeClassification:2019}
\bibfield{author}{\bibinfo{person}{Jordan Calandre}, \bibinfo{person}{Renaud
  P{\'{e}}teri}, {and} \bibinfo{person}{Laurent Mascarilla}.}
  \bibinfo{year}{2019}\natexlab{}.
\newblock \showarticletitle{Optical Flow Singularities for Sports Video
  Annotation: Detection of Strokes in Table Tennis}. In
  \bibinfo{booktitle}{{\em MediaEval}} {\em (\bibinfo{series}{{CEUR} Workshop
  Proceedings})}, Vol.~\bibinfo{volume}{2670}.
  \bibinfo{publisher}{CEUR-WS.org}.
\newblock


\bibitem[\protect\citeauthoryear{Carreira and Zisserman}{Carreira and
  Zisserman}{2017}]%
        {NN:I3DCarreira:2017}
\bibfield{author}{\bibinfo{person}{Jo{\~{a}}o Carreira} {and}
  \bibinfo{person}{Andrew Zisserman}.} \bibinfo{year}{2017}\natexlab{}.
\newblock \showarticletitle{Quo Vadis, Action Recognition? {A} New Model and
  the Kinetics Dataset}. In \bibinfo{booktitle}{{\em {CVPR}}}.
  \bibinfo{publisher}{{IEEE} Computer Society}, \bibinfo{pages}{4724--4733}.
\newblock


\bibitem[\protect\citeauthoryear{Crasto, Weinzaepfel, Alahari, and
  Schmid}{Crasto et~al\mbox{.}}{2019}]%
        {NN:3DCNNMimiciMotionForCLassification:2019}
\bibfield{author}{\bibinfo{person}{Nieves Crasto}, \bibinfo{person}{Philippe
  Weinzaepfel}, \bibinfo{person}{Karteek Alahari}, {and}
  \bibinfo{person}{Cordelia Schmid}.} \bibinfo{year}{2019}\natexlab{}.
\newblock \showarticletitle{{MARS:} Motion-Augmented {RGB} Stream for Action
  Recognition}. In \bibinfo{booktitle}{{\em {CVPR}}}.
  \bibinfo{publisher}{{IEEE} Computer Society}, \bibinfo{pages}{7882--7891}.
\newblock


\bibitem[\protect\citeauthoryear{Feichtenhofer, Pinz, and
  Zisserman}{Feichtenhofer et~al\mbox{.}}{2016}]%
        {NN:TwoStreamFusion:2016}
\bibfield{author}{\bibinfo{person}{Christoph Feichtenhofer},
  \bibinfo{person}{Axel Pinz}, {and} \bibinfo{person}{Andrew Zisserman}.}
  \bibinfo{year}{2016}\natexlab{}.
\newblock \showarticletitle{Convolutional Two-Stream Network Fusion for Video
  Action Recognition}. In \bibinfo{booktitle}{{\em {CVPR}}}.
  \bibinfo{publisher}{{IEEE} Computer Society}, \bibinfo{pages}{1933--1941}.
\newblock


\bibitem[\protect\citeauthoryear{Hanson and Pratt}{Hanson and Pratt}{1988}]%
        {Deep:weight_decay}
\bibfield{author}{\bibinfo{person}{Stephen~Jose Hanson} {and}
  \bibinfo{person}{Lorien~Y. Pratt}.} \bibinfo{year}{1988}\natexlab{}.
\newblock \showarticletitle{Comparing Biases for Minimal Network Construction
  with Back-Propagation}. In \bibinfo{booktitle}{{\em {NIPS}}}.
  \bibinfo{pages}{177--185}.
\newblock


\bibitem[\protect\citeauthoryear{Kim, Lee, and Yang}{Kim et~al\mbox{.}}{2007}]%
        {NN:3DCNN_first:2007}
\bibfield{author}{\bibinfo{person}{Ho~Joon Kim}, \bibinfo{person}{Joseph~S.
  Lee}, {and} \bibinfo{person}{Hyun~Seung Yang}.}
  \bibinfo{year}{2007}\natexlab{}.
\newblock \showarticletitle{Human Action Recognition Using a Modified
  Convolutional Neural Network}. In \bibinfo{booktitle}{{\em {ISNN} {(2)}}}
  {\em (\bibinfo{series}{Lecture Notes in Computer Science})},
  Vol.~\bibinfo{volume}{4492}. \bibinfo{publisher}{Springer},
  \bibinfo{pages}{715--723}.
\newblock


\bibitem[\protect\citeauthoryear{Lima, Fernandes, and Barros}{Lima
  et~al\mbox{.}}{2017}]%
        {NN:3DCNN:2017}
\bibfield{author}{\bibinfo{person}{Tiago Lima}, \bibinfo{person}{Bruno J.~T.
  Fernandes}, {and} \bibinfo{person}{Pablo V.~A. Barros}.}
  \bibinfo{year}{2017}\natexlab{}.
\newblock \showarticletitle{Human action recognition with 3D convolutional
  neural network}. In \bibinfo{booktitle}{{\em {LA-CCI}}}.
  \bibinfo{publisher}{{IEEE}}, \bibinfo{pages}{1--6}.
\newblock


\bibitem[\protect\citeauthoryear{Luvizon, Picard, and Tabia}{Luvizon
  et~al\mbox{.}}{2018}]%
        {Pose:ActionRecognitionAndPoseEstimation:2018}
\bibfield{author}{\bibinfo{person}{Diogo~C. Luvizon}, \bibinfo{person}{David
  Picard}, {and} \bibinfo{person}{Hedi Tabia}.}
  \bibinfo{year}{2018}\natexlab{}.
\newblock \showarticletitle{2D/3D Pose Estimation and Action Recognition Using
  Multitask Deep Learning}. In \bibinfo{booktitle}{{\em {CVPR}}}.
  \bibinfo{publisher}{{IEEE} Computer Society}, \bibinfo{pages}{5137--5146}.
\newblock


\bibitem[\protect\citeauthoryear{Martin}{Martin}{2020}]%
        {PeThesis2020}
\bibfield{author}{\bibinfo{person}{Pierre{-}Etienne Martin}.}
  \bibinfo{year}{2020}\natexlab{}.
\newblock {\em \bibinfo{title}{Fine-Grained Action Detection and Classification
  from Videos with Spatio-Temporal Convolutional Neural Networks. Application
  to Table Tennis. (D{\'{e}}tection et classification fines d'actions {\`{a}}
  partir de vid{\'{e}}os par r{\'{e}}seaux de neurones {\`{a}} convolutions
  spatio-temporelles. Application au tennis de table)}}.
\newblock \bibinfo{thesistype}{Ph.D. Dissertation}. \bibinfo{school}{University
  of La Rochelle, France}.
\newblock
\showURL{%
\url{https://tel.archives-ouvertes.fr/tel-03128769}}


\bibitem[\protect\citeauthoryear{Martin, Benois{-}Pineau, Mansencal,
  P{\'{e}}teri, Mascarilla, Calandre, and Morlier}{Martin
  et~al\mbox{.}}{2019b}]%
        {PeMETask:2019}
\bibfield{author}{\bibinfo{person}{Pierre{-}Etienne Martin},
  \bibinfo{person}{Jenny Benois{-}Pineau}, \bibinfo{person}{Boris Mansencal},
  \bibinfo{person}{Renaud P{\'{e}}teri}, \bibinfo{person}{Laurent Mascarilla},
  \bibinfo{person}{Jordan Calandre}, {and} \bibinfo{person}{Julien Morlier}.}
  \bibinfo{year}{2019}\natexlab{b}.
\newblock \showarticletitle{Sports Video Annotation: Detection of Strokes in
  Table Tennis Task for MediaEval 2019}. In \bibinfo{booktitle}{{\em
  MediaEval}} {\em (\bibinfo{series}{{CEUR} Workshop Proceedings})},
  Vol.~\bibinfo{volume}{2670}. \bibinfo{publisher}{CEUR-WS.org}.
\newblock


\bibitem[\protect\citeauthoryear{Martin, Benois{-}Pineau, Mansencal,
  P{\'{e}}teri, Mascarilla, Calandre, and Morlier}{Martin
  et~al\mbox{.}}{2021}]%
        {mediaeval/Martin21/task}
\bibfield{author}{\bibinfo{person}{Pierre{-}Etienne Martin},
  \bibinfo{person}{Jenny Benois{-}Pineau}, \bibinfo{person}{Boris Mansencal},
  \bibinfo{person}{Renaud P{\'{e}}teri}, \bibinfo{person}{Laurent Mascarilla},
  \bibinfo{person}{Jordan Calandre}, {and} \bibinfo{person}{Julien Morlier}.}
  \bibinfo{year}{2021}\natexlab{}.
\newblock \showarticletitle{Sports Video: Fine-Grained Action Detection and
  Classification of Table Tennis Strokes from videos for MediaEval 2021}. In
  \bibinfo{booktitle}{{\em MediaEval}} {\em (\bibinfo{series}{{CEUR} Workshop
  Proceedings})}. \bibinfo{publisher}{CEUR-WS.org}.
\newblock


\bibitem[\protect\citeauthoryear{Martin, Benois{-}Pineau, Mansencal,
  P{\'{e}}teri, and Morlier}{Martin et~al\mbox{.}}{2019a}]%
        {PeMEWork:2019}
\bibfield{author}{\bibinfo{person}{Pierre{-}Etienne Martin},
  \bibinfo{person}{Jenny Benois{-}Pineau}, \bibinfo{person}{Boris Mansencal},
  \bibinfo{person}{Renaud P{\'{e}}teri}, {and} \bibinfo{person}{Julien
  Morlier}.} \bibinfo{year}{2019}\natexlab{a}.
\newblock \showarticletitle{Siamese Spatio-Temporal Convolutional Neural
  Network for Stroke Classification in Table Tennis Games}. In
  \bibinfo{booktitle}{{\em MediaEval}} {\em (\bibinfo{series}{{CEUR} Workshop
  Proceedings})}, Vol.~\bibinfo{volume}{2670}.
  \bibinfo{publisher}{CEUR-WS.org}.
\newblock


\bibitem[\protect\citeauthoryear{Martin, Benois{-}Pineau, P{\'{e}}teri, and
  Morlier}{Martin et~al\mbox{.}}{2018}]%
        {PeCBMI:2018}
\bibfield{author}{\bibinfo{person}{Pierre{-}Etienne Martin},
  \bibinfo{person}{Jenny Benois{-}Pineau}, \bibinfo{person}{Renaud
  P{\'{e}}teri}, {and} \bibinfo{person}{Julien Morlier}.}
  \bibinfo{year}{2018}\natexlab{}.
\newblock \showarticletitle{Sport Action Recognition with Siamese
  Spatio-Temporal CNNs: Application to Table Tennis}. In
  \bibinfo{booktitle}{{\em {CBMI}}}. \bibinfo{publisher}{{IEEE}},
  \bibinfo{pages}{1--6}.
\newblock


\bibitem[\protect\citeauthoryear{Martin, Benois{-}Pineau, P{\'{e}}teri, and
  Morlier}{Martin et~al\mbox{.}}{2021}]%
        {Pe:3stream:2021}
\bibfield{author}{\bibinfo{person}{Pierre{-}Etienne Martin},
  \bibinfo{person}{Jenny Benois{-}Pineau}, \bibinfo{person}{Renaud
  P{\'{e}}teri}, {and} \bibinfo{person}{Julien Morlier}.}
  \bibinfo{year}{2021}\natexlab{}.
\newblock \showarticletitle{Three-Stream 3D/1D {CNN} for Fine-Grained Action
  Classification and Segmentation in Table Tennis}.
\newblock \bibinfo{journal}{{\em CoRR\/}}  \bibinfo{volume}{abs/2109.14306}
  (\bibinfo{year}{2021}).
\newblock
\showeprint{2109.14306}
\showURL{%
\url{https://arxiv.org/abs/2109.14306}}


\bibitem[\protect\citeauthoryear{Nesterov}{Nesterov}{2004}]%
        {Deep:Nesterov1}
\bibfield{author}{\bibinfo{person}{Yurii~E. Nesterov}.}
  \bibinfo{year}{2004}\natexlab{}.
\newblock \bibinfo{booktitle}{{\em Introductory Lectures on Convex Optimization
  - {A} Basic Course}}. \bibinfo{series}{Applied Optimization},
  Vol.~\bibinfo{volume}{87}.
\newblock \bibinfo{publisher}{Springer}.
\newblock


\bibitem[\protect\citeauthoryear{Nguyen, Nguyen, Ho, Nguyen, and Tran}{Nguyen
  et~al\mbox{.}}{2021}]%
        {mediaeval21/hcmus}
\bibfield{author}{\bibinfo{person}{Trong{-}Tung Nguyen},
  \bibinfo{person}{Thanh{-}Son Nguyen}, \bibinfo{person}{Gia{-}Bao~Dinh Ho},
  \bibinfo{person}{Hai{-}Dang Nguyen}, {and} \bibinfo{person}{Minh{-}Triet
  Tran}.} \bibinfo{year}{2021}\natexlab{}.
\newblock \showarticletitle{HCMUS at MediaEval 2021: Ensembles of Action
  Recognition Networks with Prior Knowledge for Table Tennis Strokes
  Classification Task}. In \bibinfo{booktitle}{{\em MediaEval}} {\em
  (\bibinfo{series}{{CEUR} Workshop Proceedings})}.
  \bibinfo{publisher}{CEUR-WS.org}.
\newblock


\bibitem[\protect\citeauthoryear{Qian, Yu, Liu, and Hauptmann}{Qian
  et~al\mbox{.}}{2021}]%
        {mediaeval21/transformer}
\bibfield{author}{\bibinfo{person}{Yijun Qian}, \bibinfo{person}{Lijun Yu},
  \bibinfo{person}{Wenhe Liu}, {and} \bibinfo{person}{Alexander~G. Hauptmann}.}
  \bibinfo{year}{2021}\natexlab{}.
\newblock \showarticletitle{Learning Unbiased Transformer for Long-Tail Sports
  Action Classification}. In \bibinfo{booktitle}{{\em MediaEval}} {\em
  (\bibinfo{series}{{CEUR} Workshop Proceedings})}.
  \bibinfo{publisher}{CEUR-WS.org}.
\newblock


\bibitem[\protect\citeauthoryear{Rogez, Weinzaepfel, and Schmid}{Rogez
  et~al\mbox{.}}{2020}]%
        {Pose:LCRNet++:2020}
\bibfield{author}{\bibinfo{person}{Gr{\'{e}}gory Rogez},
  \bibinfo{person}{Philippe Weinzaepfel}, {and} \bibinfo{person}{Cordelia
  Schmid}.} \bibinfo{year}{2020}\natexlab{}.
\newblock \showarticletitle{LCR-Net++: Multi-Person 2D and 3D Pose Detection in
  Natural Images}.
\newblock \bibinfo{journal}{{\em {IEEE} Trans. Pattern Anal. Mach. Intell.\/}}
  \bibinfo{volume}{42}, \bibinfo{number}{5} (\bibinfo{year}{2020}),
  \bibinfo{pages}{1146--1161}.
\newblock


\bibitem[\protect\citeauthoryear{Simonyan and Zisserman}{Simonyan and
  Zisserman}{2014}]%
        {NN:SimonyanTwoStream:2014}
\bibfield{author}{\bibinfo{person}{Karen Simonyan} {and}
  \bibinfo{person}{Andrew Zisserman}.} \bibinfo{year}{2014}\natexlab{}.
\newblock \showarticletitle{Two-Stream Convolutional Networks for Action
  Recognition in Videos}. In \bibinfo{booktitle}{{\em {NIPS}}}.
  \bibinfo{pages}{568--576}.
\newblock


\bibitem[\protect\citeauthoryear{Wang, Gao, Wang, Sun, and Liu}{Wang
  et~al\mbox{.}}{2018}]%
        {NN:TwoStream3D:2018}
\bibfield{author}{\bibinfo{person}{Xuanhan Wang}, \bibinfo{person}{Lianli Gao},
  \bibinfo{person}{Peng Wang}, \bibinfo{person}{Xiaoshuai Sun}, {and}
  \bibinfo{person}{Xianglong Liu}.} \bibinfo{year}{2018}\natexlab{}.
\newblock \showarticletitle{Two-Stream 3-D convNet Fusion for Action
  Recognition in Videos With Arbitrary Size and Length}.
\newblock \bibinfo{journal}{{\em {IEEE} Trans. Multimedia\/}}
  \bibinfo{volume}{20}, \bibinfo{number}{3} (\bibinfo{year}{2018}),
  \bibinfo{pages}{634--644}.
\newblock


\end{thebibliography}

\end{document}